\documentclass{article} % For LaTeX2e
\usepackage{iclr2024_conference,times}

\usepackage{amsmath,amsfonts,bm}
\usepackage{multirow}
\usepackage{tabularx}
\usepackage{graphicx}
\usepackage{adjustbox}
\usepackage{amsthm}
\usepackage{MnSymbol}
\usepackage{xspace}

\renewcommand{\arraystretch}{1.3}
\newcommand{\mset}[1]{\left\{\kern-.5em\left\{ #1 \right\}\kern-.5em\right\}}
\newcommand{\mmset}[1]{\{\kern-.4em\{ #1 \}\kern-.4em\}}

\newcommand{\vertiii}[1]{{\left\vert\kern-0.25ex\left\vert\kern-0.25ex\left\vert #1 
    \right\vert\kern-0.25ex\right\vert\kern-0.25ex\right\vert}}

\newcommand{\norm}[1]{\left\Vert#1\right\Vert}

\newcommand{\parr}[1]{\left (#1\right )}

\newcommand{\ip}[1]{\left \langle #1 \right \rangle }

\newcommand{\RN}[1]{%
  \textup{\uppercase\expandafter{\romannumeral#1}}%
}

\newcommand{\Real}{\mathbb R}

\newcommand{\too}{\rightarrow}

\newcommand{\one}{\mathbf{1}}
\newcommand{\zero}{\mathbf{0}}

\newcommand{\loss}{\mathrm{loss}}%_{\text{ur}}}
\newcommand{\rev}[1]{#1}

\newcommand{\eg}{{e.g.}}
\newcommand{\ie}{{i.e.}}

 \newtheorem{theorem}{Theorem}
 \newtheorem{lemma}{Lemma}

 \newtheorem{definition}{Definition}

\def\eqref#1{equation~\ref{#1}}
\def\1{\bm{1}}

\def\vzero{{\bm{0}}}
\def\vone{{\bm{1}}}
\def\vmu{{\bm{\mu}}}

\def\valpha{{\bm{\alpha}}}

\def\vb{{\bm{b}}}

\def\ve{{\bm{e}}}

\def\vt{{\bm{t}}}

\def\vy{{\bm{y}}}
\def\vz{{\bm{z}}}
\def\vec1{{\bm{1}}}

\def\mC{{\bm{C}}}

\def\mR{{\bm{R}}}

\def\mW{{\bm{W}}}
\def\mX{{\bm{X}}}
\def\mY{{\bm{Y}}}
\def\mZ{{\bm{Z}}}

\DeclareMathAlphabet{\mathsfit}{\encodingdefault}{\sfdefault}{m}{sl}
\SetMathAlphabet{\mathsfit}{bold}{\encodingdefault}{\sfdefault}{bx}{n}

\newcommand{\softmax}{\mathrm{softmax}}

\newcommand{\KL}{D_{\mathrm{KL}}}
\newcommand{\Var}{\mathrm{Var}}

\DeclareMathOperator*{\argmax}{arg\,max}
\DeclareMathOperator*{\argmin}{arg\,min}

\usepackage[utf8]{inputenc} % allow utf-8 input
\usepackage[T1]{fontenc}    % use 8-bit T1 fonts
\usepackage{hyperref}       % hyperlinks
\usepackage{url}            % simple URL typesetting
\usepackage{tabularx,booktabs}       % professional-qualit  y tables
\usepackage{amsfonts}       % blackboard math symbols
\usepackage{nicefrac}       % compact symbols for 1/2, etc.
\usepackage{microtype, xcolor}      % microtypography
\usepackage{wrapfig}
\usepackage{graphicx}
\usepackage{enumerate}  
\usepackage{multirow}
\usepackage{xfrac}

\usepackage{algpseudocode}
\usepackage{algorithm}

\title{Approximately Piecewise E(3) Equivariant Point Networks}

\author{Matan Atzmon$\, ^1$ Jiahui Huang$\, ^1$ Francis Williams$\, ^1$ Or Litany$^1$ $^2$  \\
$^1$ NVIDIA \ \
$^2$ Technion \\
\texttt{\{matzmon,jiahuih,fwilliams,olitany\}@nvidia.com}}
\iclrfinalcopy % Uncomment for camera-ready version, but NOT for submission.
\begin{document}
\newcommand{\ShortName}{APEN\xspace}

\maketitle
\begin{abstract}

Integrating a notion of symmetry into point cloud neural networks is a provably effective way to improve their generalization capability. Of particular interest are $E(3)$ equivariant point cloud networks where Euclidean transformations applied to the inputs are preserved in the outputs. Recent efforts aim to extend networks that are equivariant with respect to a single global $E(3)$ transformation, to accommodate inputs made of multiple parts, each of which exhibits local $E(3)$ symmetry. %In practical settings, however, the local partitioning points is \emph{unknown} apriori. 
In practical settings, however, the partitioning into individually transforming regions is \emph{unknown} a priori. %Thus, past work can only approximate piecewise symmetry due to prediction error in the partition of the points. 
Errors in the partition prediction would unavoidably map to errors in respecting the true input symmetry. Past works have proposed different ways to predict the partition, which may exhibit uncontrolled errors in their ability to maintain equivariance to the actual partition. To this end, we introduce \ShortName: a general framework for constructing approximate piecewise-$E(3)$ equivariant point networks. Our framework offers an adaptable design to  \emph{guaranteed} bounds on the resulting piecewise $E(3)$ equivariance approximation errors. 
%
%Our principal insight is that the model class of piecewise equivariant functions induced by the (unknown) true partition, contains functions that are piecewise equivariant with respect to \textit{finer} partitions. 
Our primary insight is that functions which are equivariant with respect to a \textit{finer} partition (compared to the unknown true partition) will also maintain equivariance in relation to the true partition. Leveraging this observation, we propose a compositional design for a partition prediction model. It initiates with a fine partition and incrementally transitions towards a coarser subpartition of the true one, consistently maintaining piecewise equivariance in relation to the current partition.
% We leverage this fact to suggest a partition prediction model, for which 
As a result, the equivariance approximation error can be bounded solely in terms of (i) uncertainty quantification of the partition prediction, and (ii) bounds on the probability of failing to suggest a proper subpartition of the ground truth one. %To mitigate expressivity issues resulting from considering finer partitions, we introduce a compositional design of piecewise equivariant layers that gradually consider coarser partitions. -- OrL: i think we can ommit the expressivity issue from the abstract. 
We demonstrate the practical effectiveness of \ShortName using two data types exemplifying part-based symmetry: (i) real-world scans of room scenes containing multiple furniture-type objects; and, (ii) human motions, characterized by articulated parts exhibiting rigid movement. Our empirical results demonstrate the advantage of integrating piecewise $E(3)$ symmetry into network design, showing a distinct improvement in generalization accuracy compared to prior works for both classification and segmentation tasks.

\end{abstract}

\vspace{-7pt}
\section{Introduction}
In recent years, there has been an ongoing research effort on the modeling of neural networks for 3D recognition tasks.
Point clouds, as a simple and prevalent 3D input representation, have received substantial focus, leading to \emph{point networks}: specialized neural network architectures operating on point clouds \citep{qi2017pointnet,zaheer2017deep}.
Since many point cloud recognition tasks can be characterized as equivariant functions, modeling them with an equivariant point network has been shown to be an effective approach. Indeed, equivariant modeling can simplify a learning problem: knowledge learned from one input, automatically propagates to all input's symmetries\citep{ bietti2021sample,elesedy2021provably,tahmasebi2023exact}.

One important symmetry exhibited in point clouds is the \emph{Euclidean motions}, $E(3)$, consisting of all the possible rigid motions in space. Building on the demonstrated success of $E(3)$ equivariant point networks in prior research \citep{thomas2018tensor}, recent efforts have been dedicated to extending $E(3)$ symmetry to model \emph{piecewise} rigid motions symmetry as well \citep{yu2022eon,lei2023efem,deng2023banana}. This extension is valuable since some recognition tasks can be better characterized as piecewise $E(3)$ equivariant functions. To support this claim, we turn to the task of instance segmentation within a scene, illustrated by a 2D toy example in the right inset.
\begin{wrapfigure}[5]{r}{0.33\textwidth}
  \begin{center}
  \vspace{-15pt}
    \includegraphics[width=0.33\textwidth]{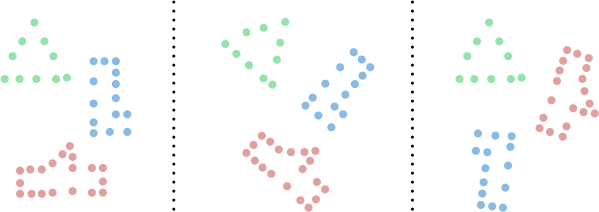}
  % \vspace{20pt}
  \end{center}
  \label{fig:inset_intro}
\end{wrapfigure}
In the leftmost column, we visualize segmentation predictions by distinct colors. In the middle column, we observe the expected invariant predictions under a global Euclidean motion of the entire scene. Finally, in the right column, we showcase invariant predictions under a \emph{piecewise} deformation that allows individual objects to move independently in a rigid manner, decoupled from the overall scene's motion.
% While a simple $E(3)$ invariant network can model the transformation from the left to the middle panel, it \emph{cannot} model the piecewise rigid transformation from the left to the right panel.
% In addition to scenes, piecewise $E(3)$ symmetry can also be applicable to other types of modalities such as articulated objects, \eg, human body. 
% Thus, incorporating piecewise $E(3)$ deformations into equivariant network design has the potential to enhance the network's generalization capacity across a variety of practical applications.  
% \orl{Note that while the DOF of piecewise $E(3)$ grows linearly with the number of parts, the number of actual configurations scales exponentially. }

% why is it hard?
Incorporating piecewise $E(3)$ symmetry to point networks presents several challenges. The primary hurdle is the unknown partitioning of the input point cloud into its moving parts.  
While having such a partition makes it possible to implement equivariant design using a $E(3)$ equivariant \emph{siamese} network across parts \citep{atzmon2022frame}, this is often infeasible in real-world applications. For instance, in the segmentation task shown in the inset, the partition is inherently tied to the model's segmentation predictions. Thus, in cases where the underlying partition is not predefined but rather predicted by a (non-degenerated) model, any suggested piecewise equivariant model will introduce an \emph{approximation error} in satisfying the equivariance constraint. \rev{We will use the term \emph{equivariance approximation error} to refer to the error that arises when a function is unable to satisfy the piecewise $E(3)$ equivariance constraint (w.r.t. the true unknown partition); see Definition \ref{def:p_equiv_def}.}%; see definition \ref{def:p_equiv_def}}.  
This equivariance approximation error is inherent unless the partition prediction remains perfectly consistent under the input symmetries. This implies it must be invariant to the very partition it seeks to identify. So far in the literature, less attention has been given to piecewise equivariant network designs that offer means to control the network's equivariance approximation error. For example, \cite{liu2023selfsupervised} suggests an initial partition prediction model based on input points' global $E(3)$ invariant and equivariant features. In \cite{yu2022eon}, local-context invariant features are used for the partition prediction model. In both cases, it is unclear how failures in the underlying partition prediction model will affect the equivariance approximation error. Notably, the concurrent work of \cite{deng2023banana} also observes the equivariance approximation error. Their work suggests an \emph{optimization-based} partition prediction model based on (approximately) contractive steps, striving to achieve exact piecewise equivariance; errors in the partition prediction model arising from expanding steps and their impact on the resulting equivariance approximation error are not discussed. 
% Another example is the work of \cite{deng2023banana}, where the partition prediction model is based on an approximate-contractive iterative process. In both cases, it is not clear how \emph{uncertainty} in the partition prediction model will affect the resulting piecewise equivarince approximation error. 

% In other words, we can't rely on a partitioning in the input since this partition is precisely what is being predicted by the network \citep{deng2023banana}. \orl{what are other challenges?}

% In this paper, we propose a novel framework for the design of piecewise equivariant networks, called \ShortName. Our method alleviates the need for input partition information for equivariant predictions. \ShortName provides a practical construction for point networks tackling basic 3D recognition tasks such as classification and segmentation. Our framework is built on the fact that any piecewise (in/equi)variant function can be partitioned into smaller pieces which are also (in/equi)variant. For example, rotating an object by 90 degrees is equivalent to rotating two halves of that object by 90 degrees.

In this paper, we propose a novel framework for the design of approximately piecewise equivariant networks, called \ShortName. Our goal is to suggest a practical design that can serve as a backbone for piecewise $E(3)$ equivariant tasks, while identifying how elements in the design control the piecewise equivariance approximation error. 
% Our method
% alleviates the need for input partition information for equivariant predictions. \ShortName provides a practical construction for point networks tackling basic 3D recognition tasks such as classification and segmentation
Our framework is built on the following simple fact. Let $G$ and $G'$ be two symmetry groups for which each symmetry in $G'$ is also in $G$, \ie, $G'\leqslant G$. Then, any $G$ equivariant function is also a $G'$ equivariant function.
% That is, if $\mathcal{F}_G = \left\{\phi:U\rightarrow V  \vert \phi \text{ is } G \text{ equivariant}\right\}$, where $U,V$ are some vector spaces, then
% \begin{equation}
%     G' \leqslant G \implies \mathcal{F}_G \subseteq \mathcal{F}_{G'}.
% \end{equation} \fran{I wonder if we can make this a high level description and push the mathematical insight into the method section?}
Thus, we can have an \emph{exact} piecewise equivariant model, as long as the model partition is a proper subpartition of the (unknown) ground-truth one.
\begin{wrapfigure}[5]{r}{0.23\textwidth}
  \begin{center}
  \vspace{-16pt}
    \includegraphics[width=0.23\textwidth]{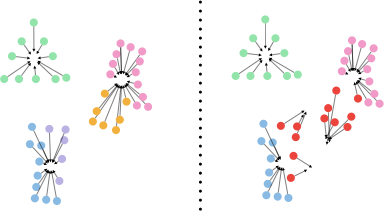}
  % \vspace{20pt}
  \end{center}
  \label{fig:inset_intro}
\end{wrapfigure}
The right inset illustrates this fact: the piecewise equivariant predictions of vote targets, marked as black dots, are accurate for a subpartition of the ground-truth partition (left column), whereas an equivariant approximation error arises for a partition that includes a \rev{\emph{bad}} part consisting of points mixed from two different parts in the ground truth partition (\ie, the red dots in the right column). 
This observation may lead to the following \emph{simple} model for partition prediction -- drawing a random partition from the distribution of non-degenerated partitions of size $k$ (\ie, all $k$ parts get at least one point). \rev{For such a model, the probability of drawing a bad part reaches $0$ as $k$ increases. In turn, the probability of drawing a bad partition can be used to bound the equivariance approximation error of a piecewise equivariant function, as good sub-partitions induce \emph{no} equivariance approximation error. Importantly, this approach alleviates the need for additional constraints on the underlying model function to control the equivariance approximation error.} 

% such as restricting the model function bounded variation}. This property is especially useful for neural networks,  which in general, bounding their variation is untractable to their design \citep{szegedy2013intriguing,anil2019sorting}.

% Therefore, this simple model enables bounding the equivariance approximation error \emph{solely} by the probability of drawing a ``bad'' partition. Crucially, this bound is independent of any required restriction on the resulting piececwise equivariant model function \emph{bounded variation}. 

% Thus, we can define a $G$ equivariant model class by adopting an \emph{over parametrization} approach which includes many more deformations than the minimally needed, with the aim that these also include the unknown piecewise $E(3)$ deformation.

However, this approach needs to be pursued with caution, as increasing the complexity of the possible partitions reduces the expressivity of the resulting piecewise equivariant point network model class. This caveat is especially relevant to the common design using a \emph{shared} (among parts) $E(3)$ equivariant backbone. %Indeed, by considering the limiting case, where each point belongs to a different part, the only shared backbone $E(3)$ equivariant functions are constant.
Indeed, at the limit where each point belongs to a distinct part, the only shared backbone $E(3)$ equivariant functions are constant.
% This raises the question of finding a deformation class that on one hand is expressive enough to include the underlying plausible piecewise deformations, while on the other hand, avoids excessive complexity, preventing the emergence of a \emph{degenerate} equivariant function class. 
%To mitigate the potential expressivity issues, we suggest a compositional network architecture consisting of a sequence of piecewise equivariant layers that the complexity of their underlying partition gradually reduces. 
To mitigate potential expressivity issues, our \ShortName framework employs a compositional network architecture. This architecture comprises a sequence of piecewise equivariant layers, with the complexity of their underlying partition decreasing gradually.
Each layer is defined as a piecewise $E(3)$ equivariant function, which not only predicts layer-specific features but also parametrizes a prediction of a \emph{coarser} partition. This coarser partition serves as the basis for the subsequent layer's piecewise $E(3)$ symmetry group. The goal of this ``bottom-up'' approach is to allow the network to overcome the issue of ambiguous predictions in earlier layers by learning to merge parts that are likely to transform together, resulting in a simpler partition in the subsequent equivariant layer. Importantly, this design also provides bounds for the piecewise equivariant approximation error of each layer, resulting solely from two sources in the design: (i) uncertainty in the partition prediction model, and (ii) the probability of drawing a bad partition. 

% At first glance, it may seem that we have replaced our initial challenge with another challenge of equivalent difficulty: learning over a complex discrete structure representing the potential partitions an input may have.
% However, our key insight is that this challenge can be mitigated by parametrizing the partitions as an outcome of layer predictions in the geometrical space, $\Real^3$. Critically, this approach enables geometrical information to be used for the partition prediction of intermediate layers, and provides a simpler and continuous domain for learning.
% \jia{Not sure if this explanation is a critical motivation? Or it just over-complicates the logic?}

We instantiated our \ShortName framework for two different recognition tasks: classification and part segmentation. We conducted experiments using datasets comprising of (i)  articulated objects consisting of human subjects performing various sequence movements \citep{dfaust:CVPR:2017}, and (ii) real-world room scans of furniture-type objects \citep{huang2021multibodysync}. The results validate the efficacy of our framework and support the notion of potential benefits in incorporating piecewise $E(3)$ deformations to point networks.

\section{Method}

\subsection{Background: Equivariant Point Networks}
We will consider point networks as functions $h: U \rightarrow W$, where $U$ and $W$ denote the vector spaces for the input and output domains, respectively. The input vector space $U$ takes the form $U = \Real^{n\times (2\times d)}$, with $n$ denoting the number of points in the input point cloud, $d$ is the point embedding space dimension (usually $d=3$), and $2$ per-point features: spatial location and an \emph{oriented normal} vector. Depending on the task at hand, classification, or segmentation, the output vector space $W$ can be $W = \Real^c$ or $W = \Real^{n\times c}$. To incorporate symmetries into a point network, we consider a group $G$ , along with its action $g$ on the vector spaces $U$ and $W$. Of particular interest in our work is the Euclidean motions group $G = E(d)$ defined by rotations, reflections and translations in $d$-dimensional space. The group action on $\mX \in U$ is defined by $g\cdot \mX = \mX \mR^T + \vone \vt^T$, with $g = (\mR,\vt)$ being an element in $E(d)$\footnote{Note that in fact $g=(\mR,\zero)$ on the input normals features.}, while the action on the output $\mY\in W$ varies depending on the task (\eg, $g\cdot \mY = \mY$ for classification). 
% The action on outputs $\mY\in W$ varies depending on the task and structure of $W$: For instance, in classification and segmentation tasks, we want the action $g\cdot \mY = \mY$ \ie \emph{invariant predictions}. 
An important property for our networks $h$ to satisfy is \emph{equivariance} with respect to $G$: 
\begin{equation}
    h(g\cdot \mX) = g\cdot h(\mX) \qquad \forall g \in G, \mX \in U.
\end{equation}

We consider the typical case of networks $h$ which follow an \emph{encoder-decoder} structure, \ie, $h = \mathtt{d} \circ \mathtt{e}$. 
The encoder $\mathtt{e}: U \rightarrow V$ transforms an input into a learnable latent representation $V$. In our case, $V$ is an $E(3)$ equivariant latent space, up to order type 1, of the form $V=\Real^{a+b\times3}$, with $a,b$ being positive integers. 
% We remark that the equivariance property of $\mathtt{e}$ is crucial for preserving symmetry information during learning.
% \fran{I want to add a sentence saying that the equivariance property of the encoder is crucial to preserve information about input deformation which }
% Nevertheless, for the applications considered in this paper, $V$ includes up to order $1$ features, \ie, $V$ takes the form $V=\Real^{a+b\times3}$, with $a,b$ being positive integers. 
The decoder $\mathtt{d}: V \rightarrow W$ decodes the latent representation to produce the expected output response which can be invariant or equivariant to the input. 
% \jia{I don't think the above descriptions (from 'Typically...') are needed because they are never used afterwards. Moreover, the notation of the decoder $g$ clashes with the group $g$.}\orl{it's discussed in 2.4 when we describe the \ShortName encoder and decoder}
Both $\mathtt{e}$ and $\mathtt{d}$ are modeled as a composition of multiple invariant or equivariant layers. Having covered the basics of equivariant point networks, we will now proceed to describe our proposed framework, starting with the formulation of a piecewise $E(d)$ equivariant layer.

\subsection{Piecewise $E(d)$ Equivariance Layer}
%Our suggestd design models a piecewise $E(d)$ equivariant network with respect to a partition predicted by a .
% . As mentioned in the introduction, we would like to control the complexity of possible deformation of the underlying symmetry group. To this end, we consider a discretized deformation model, consisting solely of piecewise $E(d)$ deformations. In turn, the number of possible parts partitions and the possible points to parts assignment determine the underlying complexity. 
%
% \jia{Shall we introduce the precise definition of what is a 'piecewise' equivariance function here? We could define 'Deterministic equivariance' (Lemma 1) for 'Known input partition' and 'Stochastic equivariance' (Definition 1) for 'Unknown input partition'.}
% At the core of our design is the modeling of a piecewise $E(d)$ equivariant network with respect to a partition predicted by a model.
% Our aim is to suggest a layer whose piecewise equivariance approximation error is bounded solely by the probability that the partition prediction model failed at drawing a proper subpartition of the ground truth one.  
%Our design suggests of a piecewise $E(d)$ equivariant network with respect to a partition predicted by a model.

% Our aim is to suggest a layer whose piecewise equivariance approximation error is bounded solely by the probability that the partition prediction model failed at drawing a proper subpartition of the ground truth one.  
We start this section by describing the settings for which we model a piecewise $E(d)$ equivariant layer.
Let $\mX \in U$ be the input to the layer. Our assumption is that the partition prediction is modeled as a (conditional) probability distribution, $Q_{\mZ\vert\mX} \in (\Sigma_k)^n$ over the $k$ parts partitions $\mX$ can exhibit. Here $\Sigma_k$ denotes the $k$ probability simplex. Let $\mZ=\left[\vz_1^T,\cdots,\vz_n^T\right]^T \in \left\{0,1\right\}^{n\times k}$  with $\mZ\vone=\vone$, denote a realization of a partition from $Q_{\mZ\vert\mX}$, i.e., $\mZ \sim Q_{\mZ\vert\mX}$. 

% Given an input $\mX \in U$, we assume the predicted partition is modeled as a joint probability distribution $Q_{\mZ\vert\mX} \in (\Sigma_k)^n$ over $n$ points and $k$ partitions of $\mX$. Here $\Sigma_k$ denotes the $k$ probability simplex, and $\mZ=\left[\vz_1^T,\cdots,\vz_n^T\right]^T \in \left\{0,1\right\}^{n\times k}$  with $\mZ\vone=\vone$ is a realization of a partition from $Q_{\mZ\vert\mX}$, i.e., $\mZ \sim Q_{\mZ\vert\mX}$. 

Let $\widehat{\mZ}$ be the \emph{unknown} ground truth partition of $\mX$. An important quantity of interest is
\begin{equation}
\rev{
    \lambda(Q) = P_{\mZ \sim Q_{\mZ\vert \mX}}\left( \exists \ 1\leq i,j \leq n    \text{ s.t. } (\mZ  \mZ^T)_{ij} > (\widehat{\mZ}  \widehat{\mZ}^T)_{ij} \right)} ,
\end{equation}
measuring the probability of drawing a ``bad'' partition from $Q$, \ie, a non-proper subpartition of $\widehat{\mZ}$.
In that context, a reference partition prediction model is $Q_{\textrm{simple}}$ which is defined by a uniform draw of a partition satisfying $\vone^T \mZ \ve_j > 0$ for each $j\in[k]$. An important property of $Q_\text{simple}$ is $\lambda(Q_{\textrm{simple}}) \rightarrow 0$ as $k \rightarrow n$.
\rev{To better understand this claim about $\lambda(Q_{\text{simple}})$, one can consider the sequential process generating a random $k$ parts partition. Clearly, larger values of $k$ result in each part containing fewer points. Since the probability of drawing the next point from mixed ground-truth parts is independent of $k$, determined solely by the number of input points and the ground-truth partition, the probability that the next drawn point generated a bad part lowers as $k$ increases.} In turn, $\lambda(Q_{\textrm{simple}})$ can serve as a useful bound for the resulting equivariance approximation error. Consequently, we opt for a model $Q$ that satisfies $\limsup \lambda(Q) = \lambda(Q_\textrm{simple})$,  where the last limit is taken with respect to a hyper-parameter in the design of $Q$. 

More precisely, we suggest the following characterization for $Q$.  Let $\delta:(\Sigma_k)^n \rightarrow \Real_+$, satisfying 
\begin{equation}\label{eq:delta}
    \delta(Q) \rightarrow 0, \text{ whenever } Q \rightarrow Q_v
\end{equation}
with $Q_v \in \left\{ 0,1 \right\}^{n\times k} \cap (\Sigma_k)^n$. That is, $\delta$ measures the uncertainty in the model's prediction. Our design \emph{requirement} is that 
\begin{equation}\label{eq:lambda}
    \limsup\lambda(Q) = \lambda(Q_\textrm{simple}), \text{ as } \delta(Q) \rightarrow 0.
\end{equation}
\begin{wrapfigure}[10]{r}{0.22\textwidth}
  \begin{center}
  \vspace{-20pt}
    \includegraphics[width=0.22\textwidth]{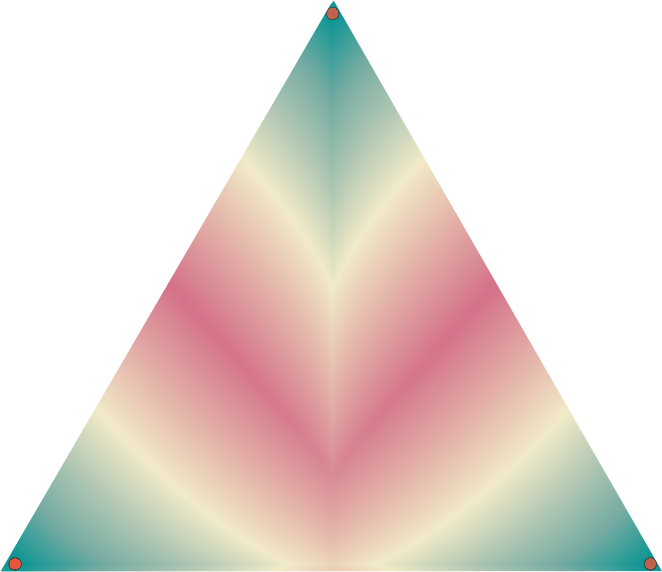}
  \end{center}
  \vspace{-15pt}
  \caption{The functional bound $\delta$. Green colors indicate values close to $0$.}\label{fig:lambda}
\end{wrapfigure}
In other words, we suggest constraining a $Q$ model to behave in a way such that, as it becomes more \emph{certain} in how it draws its predictions, the probability of drawing a ``bad'' partition converges to be \emph{no worse} than the one of the simple model. The functional $\delta$ measures the uncertainty of $Q$, and is considered as one of the design choices in the modeling of $Q$. In turn, it will be used to bound the equivariance approximation error. Fig.~\ref{fig:lambda} illustrates the qualitative behavior of $\delta$.%; green colors indicate values close to $0$.

We defer the discussion on how we provide a model for $Q$ supporting these ideas for later. Instead, we start by describing how $Q_{\mZ\vert \mX}$ is incorporated to model a piecewise equivariant layer.

% To that end, we will model that possible partitions an input my exhibit using a (conditional) random variable
\paragraph{Fixed partition.}
To facilitate discussion, we first assume that $\mZ$ is fixed, and we will start by describing a piecewise $E(d)$ equivariant layer with respect to $\mZ$ partition.
% Given a partition, we start by defining the piecewise $E(d)$ action. 
Let $G=E(d)\times\cdots\times E(d)$ be the product consisting of $k$ copies of the Euclidean motions group. For $g=(g_1\cdots,g_k)\in G$, we define 
\begin{equation}\label{eq:action}
g\cdot (\mX,\mZ) = \sum_{j=1}^{k} \left(g_j\cdot \mX\right) \odot (\mZ \ve_j \one_d^T),
\end{equation}
where $g_j \cdot \mX = \mX \mR_j^T + \one_n \vt_j^{T} $, $\left\{\ve_j\right\}_{j=1}^k$ is the standard basis in $\Real^k$, $\vone_d$ is the vector of all ones in $\Real^d$, and $\odot$ denotes the Hadamard product between two matrices.

One appealing way to model a piecewise $E(d)$ equivariant function, $\psi : U \times \left\{0,1\right\}^{n\times k}\rightarrow U'$, which also respects the inherited order symmetry of the part's assignments, is by employing an $E(d)$-equivariant backbone $\psi_b : U \rightarrow U'$ shared among the parts \citep{atzmon2022frame,deng2023banana}, taking the form: 
\begin{equation}\label{eq:pe_arch}
    \psi(\mX,\mZ) = \sum_{j=1}^k \psi_b(\mX \odot \mZ\ve_j\vone_d^T) \odot \mZ \ve_j\vone^T.
\end{equation}
The following lemma, whose proof can be found in the Appendix, verifies these properties for $\psi$.
\begin{lemma}\label{lm:basic}
    Let $\psi: U \times \left\{0,1\right\}^{n\times k} \rightarrow U'$ be a function as in Eq.~(\ref{eq:pe_arch}).
    Let $g\in G$ and $\sigma_k(\cdot)$ a permutation on $[k]$. Then,
    \begin{align*}
        \psi(g\cdot(\mX,\mZ), \mZ) &= g\cdot (\psi(\mX,\mZ),\mZ), \\
        \psi(\mX,\mZ') &= \psi(\mX,\mZ)
    \end{align*}
    for any $\mX \in U$, $\mZ \in \left\{0,1\right\}^{n\times k}$, and $\mZ' = \mZ_{:,\sigma(i)}$.
\end{lemma}
Note that one can consider augmenting the design of Eq.~(\ref{eq:pe_arch}) with a function over orderless representation of parts $E(d)$ invariant features \citep{maron2020learning}. Equipped with the construction in Eq.~(\ref{eq:pe_arch}), we will now move on to the case where $\mZ$ is uncertain. 

\paragraph{Uncertain partition.} Incorporating ${Q_{\mZ\vert\mX}}$ into a layer can be done by marginalizing over the possible $\mZ$. Some simple options for marginalization are i) $\phi_{\RN{1}}(\mX) = \psi(\mX,\mathbb{E}_Q\mZ)$ as implemented in \cite{atzmon2022frame}; ii) $\phi_{\RN{2}}(\mX) = \mathbb{E}_{Q}\psi(\mX,\mZ)$; and iii) $\phi_{\RN{3}}(\mX)=\psi\left(\mX, \mZ_{*} \right)$, where $(\mZ_{*})_{i,:} = \ve_{\argmax_{j} Q(\mZ \vert \mX)_{ij}}$. Unfortunately, however, all of these options are merely an \emph{approximation} of a piecewise $E(d)$ equivariant function. 
The scheme $\phi_{\RN{1}}$ relies on scaling, which can be an arbitrarily bad approximation to the input's geometry. The scheme $\phi_{\RN{2}}$ relies on the averaging of equivariant point \emph{features}, which is not stable under a realization of a particular partition $\mZ \sim Q$. Similarly, $\phi_\RN{3}$ is also not equivariant under all possible realizations of $\mZ$. However, the equivariance approximation error $\phi_\RN{3}$ induces can be controlled, as we discuss next. 

\paragraph{Bounding the equivariant approximation error.}In this work, we advocate for layers of the form $\phi_{\RN{3}}$. The motivation for doing so is that it enables a \emph{uniform} control over the equivariant approximation error as a function of $Q$, crucially, without relying on bounding the \emph{variation} of $\phi$. This advantage is especially prominent for neural networks, as existing techniques for bounding network's bounded variation, \eg, by controlling the network's Lipshitz constant, impose additional complexity to the network architecture and may hinder the training process \citep{anil2019sorting}.
%are elusive as they rely on the addition of loss terms or some heuristics. 
On the other hand, as we will see in the next section, the approximation error $Q$ induces can be controlled explicitly by a choice of hyper-parameters in the parametrization of $Q$.

The next definition captures our suggested characterization for an approximation error of a desired piecewise $E(d)$ equivariant layer:

\begin{definition}\label{def:p_equiv_def}
    Let $\phi:U \rightarrow U'$ be a bounded function with $\norm{\phi} \leq M$. Let $\delta:(\Sigma_k)^n \rightarrow \Real_+$, satisfying Eq.~(\ref{eq:delta}) and Eq.~(\ref{eq:lambda}) w.r.t. $Q$. Then, $\phi$ is a $(G,Q)$ equivariant function if and only if for any given $\mX \in U$, the following is satisfied
    \begin{equation}
        \mathbb{E}_{Q_{\mZ|\mX}} \norm{\phi\left( g\cdot (\mX,\mZ) \right) - g\cdot (\phi (\mX),\mZ)} \leq \left(\lambda(Q_\textrm{simple}) + \delta(Q)\right) M
    \end{equation}
    for all $g \in G$. We denote the set of $(G,Q)$ equivariant functions by $\mathcal{F}_{Q}$.
\end{definition}

The above characterization for the equivariance approximation error can be seen as resulting from two different sources of properties in the partition prediction model: (i) an intrinsic source, as captured by $\delta$, which measures the uncertainty of the model $Q$, and (ii) an extrinsic source, determined by a measure independent from $Q$ as captured by $\lambda$.
In addition, the above definition generalizes the notion of exact equivariant function classes. For instance, consider $\widehat{\mZ}$ satisfying $\widehat{\mZ} \ve_j= \vone $ for some \emph{fixed} $j$; setting $\delta \equiv 0$ yields that $\mathcal{F}_Q$ coincides with the class of global $E(d)$ equivariant functions. 
% Fig.~\ref{fig:lambda} illustrates the expected behavior of $\lambda$; green colors indicate values close to $0$. Importantly, since for some common techniques for the parametrization of $Q$, a simple choice of hyper-parameters determines how $Q$ is restricted to be close to one of the simplex vertices, then $\lambda(Q)$ is in fact a tunable bound of for the layer's model class. 

To conclude this section, we verify in the following theorem that our construction of $\phi$ indeed falls under the suggested characterization of approximate piecewise $E(d)$ equivariant functions. Proof details are in the Appendix.
\begin{theorem}\label{thm:main}
    Let $\phi:U \rightarrow U'$ be of the form 
    \begin{equation}\label{eq:arch_final}
      \phi(\mX) = \sum_{j=1}^k \psi_b(\mX \odot \mZ_{*} \ve_j \vone^T_d)\odot \mZ_{*} \ve_j \vone^T,  
    \end{equation}
    
    where $(\mZ_{*})_{i,:} = \ve_{\argmax_{j} Q(\mZ \vert \mX)_{ij}}$, and $\psi_b : U \rightarrow U'$ is an $E(d)$ equivariant backbone. Then, 
    $$\phi \in \mathcal{F}_Q.$$
\end{theorem}

\vspace{-15pt}
\subsection{Q Prediction}\label{ss:q_pred}
\vspace{-5pt}
So far, we have treated $Q$ as a given input to the layer. In fact, we suggest that $Q$ results from a piecewise equivariant prediction of a prior layer. Exceptional is the first layer, for which $Q=Q_\textrm{simple}$. Given a layer output of the form in Eq.~(\ref{eq:arch_final}), we will next describe how $Q^\textrm{pred}$ is inferred. Note that $Q$ still denotes the given input partition prediction model.
\vspace{-5pt}
\paragraph{Modeling considerations.}
% As described above, the basic requirement in our design for $Q$ is to satisfy equations \ref{eq:delta} and \ref{eq:lambda}.
% Two important aspects in our design are: 
% i) Each layer, in addition to predicting piecewise $E(d)$ equivariant features for the subsequent layers, also \emph{predicts the distribution} $Q$; 
% ii) piecewise equivariant features predictions of earlier layers are induced by finer partitions compared to those of the subsequent layers. 
As a first attempt, one might consider parametrizing $Q^{\textrm{pred}}$ as the $\softmax$ of a per-point $Q$ piecewise invariant layer prediction. However, this approach introduces several difficulties, causing it to be unfeasible. Firstly, it is unclear how to supervise $Q$ during training to predict good sub-partitions of the ground-truth partition. Secondly, network optimization could be tricky, since the domain of possible partition solutions has a high dimensional combinatorial structure, especially due to our design bias for a large number of parts in early network layers. Lastly, there is a need to model the merging of parts in the input partition to generate a coarser one. 

To address these challenges, we propose a geometric approach to model $Q^{\textrm{pred}}$. Our suggestion is to set $Q^{\textrm{pred}}$ as the assignment scores resulting from the partitioning (\ie, clustering) in $R^d$ of $Q$ piecewise \emph{equivariant} per-point predictions. 
Notably, this suggestion falls under the well-known attention layer \citep{vaswani2017attention,locatello2020object,liu2023selfsupervised} following a query, key, and value structure with $\phi(\mX)$ being the values and queries, part centers being the keys, and the prediction $Q^{\textrm{pred}}$ is proportional to the matching score of a query to a key. One of the advantages of this approach is that $Q^{\textrm{pred}}$ emerges as an \emph{orderless} prediction with respect to possible parts assignments, thus simplifying the optimization domain. However, it is not clear how this model can (i) control the resulting $\delta(Q^{\textrm{pred}})$ by means of its design; and (ii) support the merging of parts to constitute a prediction of a coarser partition. To this end, we suggest that the part center (keys) predictions are set as the minimizers of an energy that is invariant to $Q$ piecewise $E(d)$ deformations of $\phi(\mX)$ (values). We formalize this idea in the next paragraph.
\vspace{-5pt}
\paragraph{Q Prediction.}
Let $\mY = [\vy_1, \cdots ,\vy_n]^T \in \Real ^ {n \times d}$ denote the first equivariant per-point prediction in $\phi(\mX) \in U'$. Let $\left[\vmu^*_j\right]_{j=1} ^k \in \Real^{d\times k} $ denote the underlying predicted part centers with which the score of $Q^\textrm{pred}$ is defined. We define $\left[\vmu^*_j\right]_{j=1} ^k$ as the minimizers of an energy consisting of the negative log-likelihood of a Gaussian Mixture Model and a regularization term that constraints the KL distance between all pairs of Gaussians to be greater than some threshold. 
Let $P(\mY; \valpha=(\vmu_j,\pi_j;\sigma)_{j=1} ^ k)$ denote the mixture distribution, parametrized by $\valpha$.
Then, the log-likelihood is $\log P(\mY; \valpha) = \sum_{i=1}^n \log(\sum_{j=1}^k \pi_j \mathcal{N}\left(\vy_i; \vmu_j, \sigma)\right)$ where $\mathcal{N}(\cdot;\vmu_j,\sigma)$ denotes the density of an isotropic Gaussian random variable, centered at $\vmu_j$ with variance $\sigma^2 I$. Note that $\sigma$ is fixed and is considered as a hyper-parameter. Then, $\left[\vmu^*_j\right]_{j=1} ^k$ are defined as
\begin{equation}\label{eq:em_min}
     (\vmu^*_j,\pi^*_j) = \argmin_\valpha -\log P(\mY; \valpha) - \tau \sum_{j\neq j'} \pi_j \pi_j' \log \KL(\mathcal{N}(\cdot;\vmu_j)\vert \vert \mathcal{N}(\cdot;\vmu_j')).
\end{equation}
In turn, the prediction $Q^{\textrm{pred}}_{ij}$ is defined as
\begin{equation}\label{eq:q_pred}
    Q^{\textrm{pred}}_{ij} = \frac{\mathcal{N}(\vy_i;\vmu^*_j,\sigma)\pi^*_j}{\sum_{j=1}^{k}\mathcal{N}(\vy_i;\vmu^*_j,\sigma)\pi^*_j}.
\end{equation}
Importantly, the above construction yields that as $\sigma \rightarrow 0$: 
i) $\lambda(Q^{\textrm{pred}}) \rightarrow \lambda(Q_{\textrm{simple}})$ since each random partition is a minimizer of the likelihood functional, and ii) $\delta(Q^{\textrm{pred}}) \rightarrow 0$.
In addition, $\sigma$ also controls the sensitivity of Gaussians to merge (under a fixed coefficient $\tau)$, where larger values encourage Gaussians to explain wider distribution of values $\vy_i$. Thus, setting an increasing sequence of $\sigma$ values across layers supports the gradual coarsening of partitions design. Lastly, note that differentiating the prediction of $Q^{\textrm{pred}}$ w.r.t. its inputs is not trivial; these details are covered in the next section.

\vspace{-5pt}
\subsection{Implementation Details}\label{ss:imp_det}
%\vspace{-5pt}
\paragraph{Network architecture.}
We start by sharing the details about the construction of the layer $\psi$ in Eq.~(\ref{eq:pe_arch}) given a known partition $\mZ$. For that end, we used Frame Averaging (FA) \citep{puny2022frame} with a shared pointnet \citep{qi2017pointnet} network, $\tilde{\psi}$. We define our shared equivariant backbone by 
$$
    \psi_b(\mX \odot \mZ \ve_j \vone^T_d) = \ip{\tilde{\psi}(\mX \odot \mZ \ve_j \vone^T_d)}_{F(\mX \odot \mZ \ve_j \vone^T_d)}
$$
where $F(\mX \odot \mZ \ve_j \vone^T_d)$ is the same PCA based construction for an $E(d)$ frame suggested in \cite{puny2022frame}, and $\ip{\cdot}$ is the FA symmetrization operator. Then, $\psi(\mX,\mZ)$ is defined exactly as in Eq.~(\ref{eq:pe_arch}). Since this construction needs to support layers with a relatively large number of parts $k$, we implement the network $\psi_b$ using the sparse linear layers from  \cite{choy20194d}.
\begin{wrapfigure}[6]{r}{0.33\textwidth}
  \begin{center}
  \vspace{-15pt}
    \includegraphics[width=0.32\textwidth]{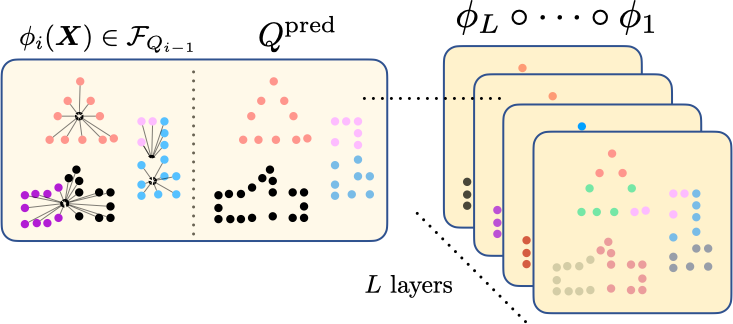}
  \end{center}
  \vspace{-15pt}
  \caption{\ShortName network design.}\label{fig:arch}
\end{wrapfigure}
In all our experiments, we implemented the encoder as a composition of $L$ layers, $\textit{e} = \phi_L \circ \cdots \circ \phi_1$, with $L = 4$; see Fig.~\ref{fig:arch}.  $Q_{\textrm{simple}}$ is set as the input to $\phi_1$. In fact, $Q_{\textrm{simple}}$ can be further regulated than the naive suggestion. In practice, we set $Q_{\textrm{simple}}$ by a Voronoi partition resulting from $k$ furthest point samples from the input $\mX$. The exact analysis of $\lambda(Q_{\textrm{simple}})$ as a function of $n$ and $k$ is out of the scope of this work -- we only rely on Eq.~(\ref{eq:lambda}).

\paragraph{Q prediction.}
  
For finding a minimizer of Eq.~(\ref{eq:em_min}), we used a slight modification of the well-known EM algorithm \citep{dempster1977maximum} that supports the merging of centers closer than the threshold $\tau$.
Note that during training, the \emph{backward} calculation requires the derivative of $\frac{\partial{Q}}{\partial \phi}$. Since the EM is in an iterative algorithm, this might unnecessarily increase the computational graph of the backward computation. To mitigate this, we use the following construction, based on implicit differentiation \citep{atzmon2019controlling,bai2019deep}. Let $\tilde{\valpha}$ be a minimizer Eq.~(\ref{eq:em_min}) that is detached from the computational graph and $\mY$. Then, $s(\mY;\tilde{\valpha}) = 0$ where $s(\mY;\tilde{\valpha}) = \nabla_{\valpha} \log P(\mY;\valpha)$, known in the literature as the score function \citep{bishop2006pattern}. We define
\begin{equation}
    \valpha = \tilde{\valpha} + I^{-1}\left(\tilde{\valpha}\right) s\left(\mY;\tilde{\valpha}\right),
\end{equation}
where $I^{-1}(\tilde{\valpha}) = \Var \left(s(\mY;\tilde{\valpha})\right)$ is the fisher information matrix \citep{bishop2006pattern} calculated at $\tilde{\valpha}$. 
Importantly, $I$ only depends on $s$ and does not involve second derivative calculations. It can be easily verified that $\valpha$ is a minimizer of Eq.~(\ref{eq:em_min}) and that  $\frac{\partial{\valpha}}{\partial{\mY}} = \frac{\partial{\left(\argmin_{\valpha}(E(\valpha,\mY))\right)}}{\partial{\mY}}$, where $E(\cdot)$ denotes the energy defined in Eq.~(\ref{eq:em_min}).
This is summarized in Alg.~\ref{alg:qpred}, found in the Appendix.

\vspace{-5pt}
\paragraph{Training details.}
Our framework requires supervision in order to train $Q^{\textrm{pred}}$ to approximate the ground-truth partition. To that end, we compute the ground-truth $\mY_{\textrm{GT}}\in \Real^{n\times d}$ to supervise the parts center \rev{vote} predictions $\mY_{l}\in \Real^{n\times d}$ \rev{of the $l^{\text{th}}$ layer}. 
We utilize the given segmentation information, to calculate $\mY_{\textrm{GT}} = \mZ \mC^T - \mX$, where $\mZ\in \left\{0,1\right\}^{n \times k}$ are the ground-truth assignments of $\mX \in \Real^{n\times d}$ and $\mC \in \Real^{d\times k}$ is calculated as the center of the minimal bounding box encompassing each of the input parts. Then, a standard $L_1$ loss is added to optimization,
$$
    \loss_{\textrm{A}} = \sum_{l=1}^L \norm{\mY_l - \mY_{\textrm{GT}}}.
$$

%More details about the selected hyper-parameters, can be found in the Appendix.

%In all of our experiments, the number of layers, $L$, was set to $4$. The choice for the sequence of $\sigma_l$ is $(0.002,0.005,0.008,0.01)$. During training
\vspace{-5pt}
\section{Experiments}
%\vspace{-5pt}
\begin{figure}[b!]
    \centering
    \includegraphics[width=\linewidth]{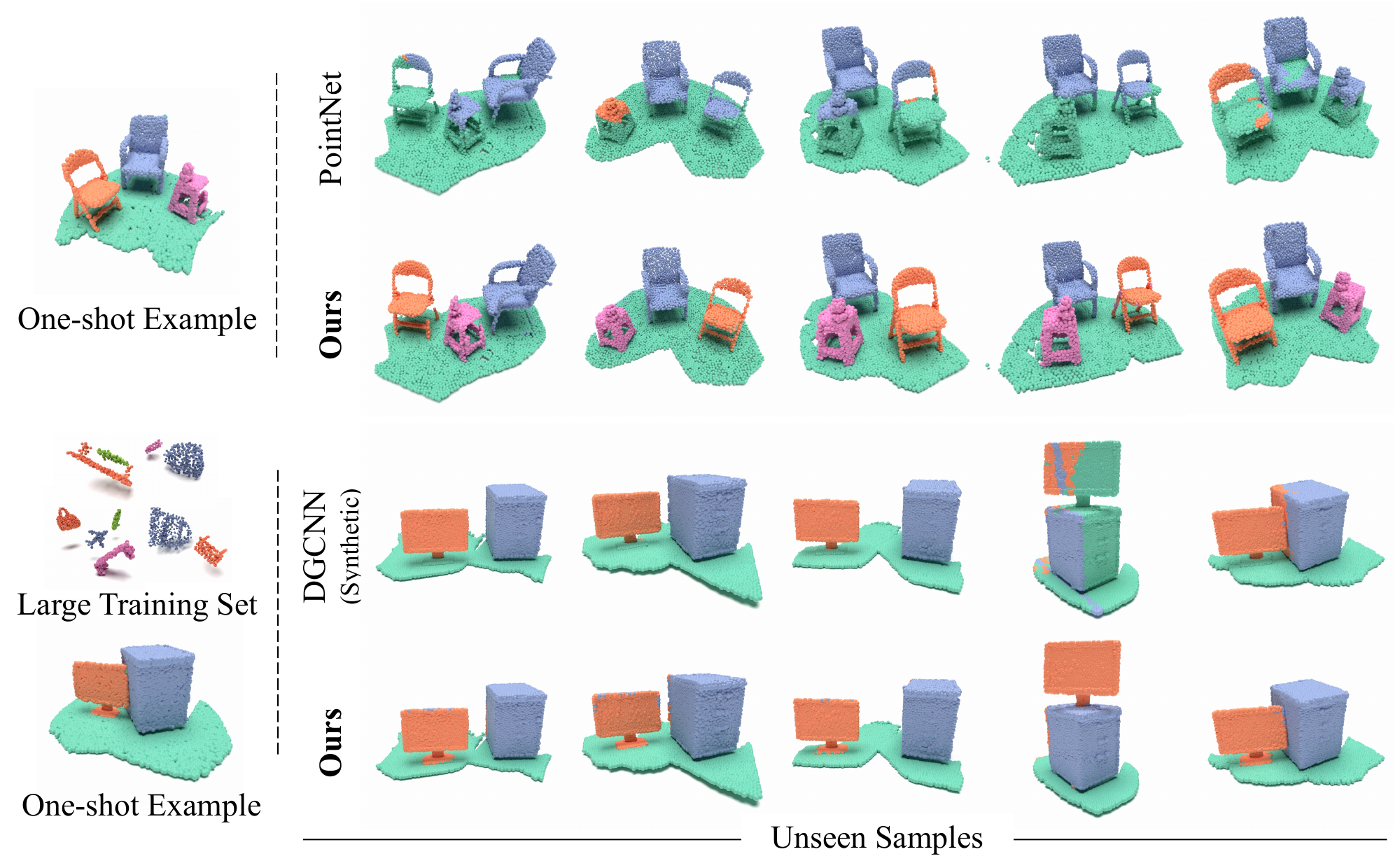}
    \caption{Qualitative results for one-shot generalization on DynLab dataset \citep{huang2021multibodysync}.}
    \vspace{-20pt}
    \label{fig:dynlab-result}
\end{figure}

We evaluate our method on two types of datasets that fit piecewise $E(3)$ symmetry: (i) scans of human subjects performing various sequences of movements \citep{SMPL:2015,dfaust:CVPR:2017,mahmood2019amass}, and (ii) real-world rooms scans of furniture-type objects \citep{huang2021multibodysync}. 
In all of our experiments, we used the ground-truth segmentation maps to extract $\mY_{\textrm{GT}}$ supervision as described in Sec.~\ref{ss:imp_det}.
\subsection{Human Scans}\label{ss:humans}

We start by evaluating our framework for the task of point part segmentation, a basic computer-vision task with many downstream applications. Specifically, we consider human body parts segmentation, where the goal is to assign each of the input scan points to a part chosen from a predefined list. In our case, the list consists of $24$ body parts. %,\eg, left hand, right hand, head, etc. 
\begin{wrapfigure}[15]{r}{0.5\textwidth}
\vspace{-20pt}
  \begin{center}
    \includegraphics[width=0.48\textwidth]{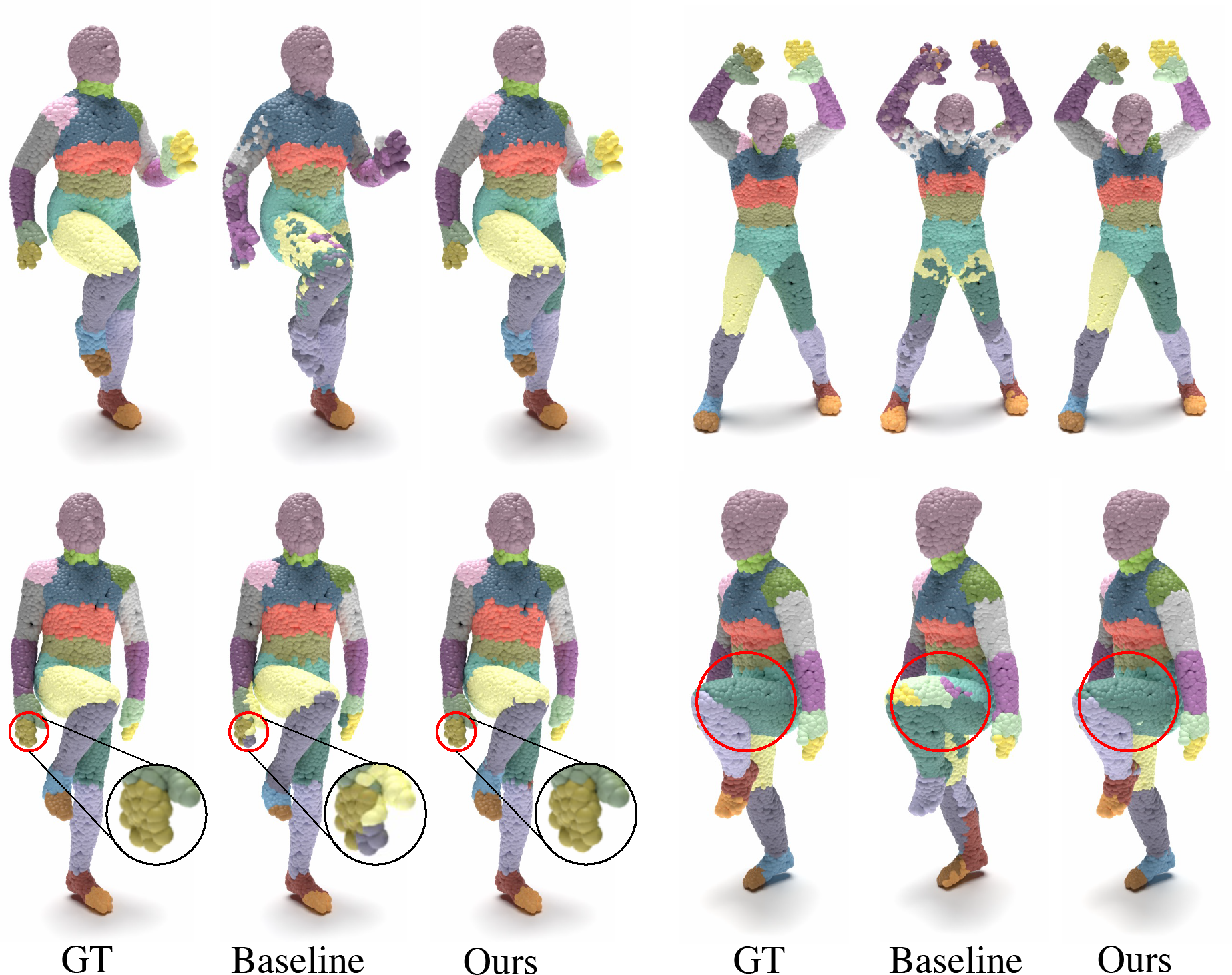}
  \end{center}
  \vspace{-15pt}
  \caption{Human body part segmentation.} %Each triplet shows ground truth, baseline, and our result in order. Baseline of the first row is \cite{deng2021vector} and the second row is \cite{qi2017pointnet}.\orl{which setting is depicted? first,second or third?}}
  \label{fig:dfaust}
\end{wrapfigure}
To evaluate different aspects of our framework, we use three different train/test splits. The first consists of  
a random ($90\%/10\%$) train/test split of $41,461$ human models from the SMPL dataset \citep{SMPL:2015} consisting of $10$ different human subjects as in \citep{huang2021arapreg}. This experiment acts as a sanity test and ensure our method does not underperform compared to baselines. The second and third splits use the scans from the Dynamic FAUST (DFAUST) dataset \citep{dfaust:CVPR:2017}, consisting of $10$ to $12$ different sequences of motions (\eg, jumping jacks, punching, etc.) for each of the $10$ human subjects.
In the second split, we divide the data by a random choice of a \emph{different} action sequences for each human. This experiment ensures our method can generalize knowledge of action sequences seen in training from one human subject to other human subjects at test time. Finally, in the third split we choose the \emph{same} sequence of movements (\eg, the one-leg jump sequence) to be removed from the training set and be placed as the test set. The last test evaluates the effect of the piecewise $E(3)$ prior, as implemented in our method, to generalize to unseen movement.

% The first test is sort of a \emph{sanity} check, \ie, evaluates that our method does not under-perform vs. baselines in settings where the train and test set are similar with respect to the distribution of subjects movements. To that end, we use a random ($90\%/10\%$) train/test split of $41,461$ human models as in \cite{huang2021arapreg}. The random models were generated using the SMPL model \cite{SMPL:2015}; in total, the dataset consists of $10$ different human subjects. For the second and third tests, we use the scans from the DFaust dataset \citep{dfaust:CVPR:2017}, consisting of $10$ to $12$ different sequences of motions for each of the $10$ human subject, \eg, jumping jacks, punching, etc.  In the second test, we set a test set by a random choice of a \emph{different} action sequence for each human. This test checks the alternative hypothesis to our method: baseline networks can generalize knowledge of action sequences \emph{seen} during training from one human subjects to other subjects at test time. Lastly, for the third test, we choose the \emph{same} sequence of movements, \eg, the one-leg jump sequence, to be removed from the training set and be placed as the test set. The last test evaluates the effect of the piecewise $E(3)$ prior, as implemented in our method, to generalize to unseen movement data.
\begin{wraptable}[10]{r}{0.45\textwidth}
    \vspace{-13pt}
    \centering
    \scriptsize
    \setlength\tabcolsep{3.8pt} % default value: 6pt
    \begin{tabular}{c}
        \begin{adjustbox}{max width=\textwidth}
            \aboverulesep=0ex
            \belowrulesep=0ex
            \renewcommand{\arraystretch}{1.2}
            \begin{tabular}{l|c|c|c}
            Method &  random & unseen random seq. & unseen seq.
                 \\ 
                    \hline
            % PointNet  & $84.4$ & $75.6$ & $78.7$  \\
            % DGCNN & $82.2$ & $60.0$ & $74.4$ \\
            % VN & $42.4$ & $15.0$ & $28.0$ \\
            PointNet  & $84.4$ & $78.5$ & $80.1$  \\
            DGCNN & $82.2$ & $70.3$ & $79.5$ \\
            VN & $42.4$ & $24.8$ & $33.3$ \\
            \rev{VN-T} & $63.5$ & $50.9$ & $50.0$ \\
            \rev{FA} & $83.5$ & $78.1$ & $76.7$ \\
            \rev{EPN} & $89.6$ & $77.8$ & $84.1$ \\
            Ours  & $\textbf{94.2}$ & $\textbf{92.2}$ & $\textbf{93.5}$
            \end{tabular}
            \end{adjustbox}
\end{tabular}
\vspace{-5pt}
\caption{ Mean IoU(\%) test set score for human body parts segmentation. \vspace{-7pt}}
\label{tab:dfaust}
\end{wraptable}
In Tab.~\ref{tab:dfaust}, we report the mean IoU(\%) score for all $3$ tests. \rev{As baseline models, we opt for PointNet \citep{qi2017pointnet} and DGCNN \citep{wang2019dynamic} as order invariant point networks. For $E(3)$ invariant networks, our baselines selection includes Vector Neurons (VN) \citep{deng2021vector}, VN-Transformer (VN-T) \citep{assaad2023vntransformer}, FrameAveraging (FA) \citep{puny2022frame}, and Equivariant Point Network (EPN) \citep{chen2021equivariant} backbone as implemented in the human body part segmentation network described in \cite{feng2023generalizing}}. Fig.~\ref{fig:dfaust} shows qualitative test results of an unseen random seq. pose (first row) and an unseen random pose (second row).
We conclude from the results that (i) our framework is a valid backbone with similar expressive power as common point network baselines, (ii) our framework utilizes piecewise $E(3)$ equivariance to gain better generalization across human subjects than baseline approaches and, (iii) piecewise $E(3)$ equivariant prior can help to generalize to unseen movements. 

Lastly, to test the versatility of our framework, we evaluate it on a point cloud classification task. On that hand, we consider the DFaust subset of AMASS \citep{mahmood2019amass}, consisting of $9$ human subjects. We define the task of classifying a model to a subject. For testing, we use an "out of distribution" test set from PosePrior \citep{akhter2015pose}. The results from this experiment support the usability of our framework for classification tasks as well. The detailed report can be found in the Appendix, including all the hyper-parameters used for the experiments in this section.

\subsection{Room Scans}\label{ss:rooms}
In this section, we test the potential of our framework for \emph{one-shot} generalization. To that end, we employ a dataset of $8$ scenes capturing a real-world room where the furniture in the room has been positioned differently in each of the 8 scans for each scene. Within each scan, there are $3$ to $4$ labeled furniture-type objects, including the floor.
% , captured at different time stamps \citep{huang2021multibodysync}. 
% These room scans include $3$ to $4$ movable furniture-type objects, including the floor. 
The task objective is to assign each input point to one of the object instances composing the scene. In addition to the difficulty of segmenting moving objects in the scene, solutions to this task must handle noise and sampling artifacts arising from the scanning procedure. For instance, scans of objects occasionally contain holes or exhibit ghost geometry. Here we compare two alternative solutions this this task: (1) we only train our method using a \emph{single} scan, and test its generalization to the other seven scans of the same scene. (2) We train baseline networks on the large-scale synthetic shape segmentation dataset from \cite{huang2021multibodysync}, which randomly samples independent motions for multiple objects taken from ShapeNet \citep{chang2015shapenet}.

In Tab.~\ref{tab:dynlab} we report the mean IoU(\%) test score for each of the scenes. Fig.~\ref{fig:dynlab-result} shows qualitative results for $2$ rooms. Despite only training on a single scan, our model outperforms baselines trained on a large synthetic dataset in 7 out of the 8 test scenes. These results suggest potential advantages of using piecewise $E(3)$ equivariant architectures in a single shot setting over the use of large-scale synthetic data.
Furthermore, to make baseline approaches work, we employed a RANSAC algorithm to identify the ground plane, with an inlier distance threshold of $0.02$ and $1000$ RANSAC iterations. In contrast, our method requires \emph{no} preprocessing since the network can treat the floor as it would for any other part of the input data.

\vspace{-6pt}
\begin{table}[h]
\centering
\resizebox{\textwidth}{!}{%
\begin{tabular}{l|llllllll}
Method &  Scene $1$ & Scene $2$ & Scene $3$ & Scene $4$ & Scene $5$ & Scene $6$ & Scene $7$ & Scene $8$ 
     \\ 
     \hline
PointNet  & $33.0\pm8.5$ & $50.2\pm4.3$ & $31.1\pm3.4$ & $38.3\pm4.3$ & $36.7\pm6.1$ & $45.2\pm22.0$ & $57.4\pm1.5$ & $36.6\pm4.6$  \\
      
DGCNN  & $36.7\pm3.6$ & $38.8\pm10.8$ & $41.8\pm4.9$ & $31.0\pm2.7$ & $48.9\pm4.3$ & $35.1\pm8.4$ & $59.5\pm7.3$ & $35.4\pm6.3$  \\

VN & $13.0\pm2.8$ & $18.6\pm1.5$ & $24.7\pm0.8$ & $15.2\pm1.1$ & $24.4\pm1.1$ & $17.6\pm1.7$ & $25.6\pm1.0$ & $23.0\pm1.2$  \\

Ours  & $\textbf{88.0}\pm13.0$ & $\textbf{98.2}\pm0.7$ & $\textbf{97.4}\pm1.5$ & $\textbf{96.3}\pm2.0$ & $\textbf{93.2}\pm3.9$ & $93.4\pm2.8$ & $\textbf{83.3}\pm13.3$ & $\textbf{92.2}\pm1.8$  \\

\hline

PointNet (Synthteic)  & $76.6\pm22.4$ & $97.3\pm2.1$ & $91.2\pm4.8$ & $89.7\pm4.0$ & $91.9\pm5.1$ & $95.1\pm1.0$ & $66.6\pm9.7$ & $83.2\pm4.0$  \\

DGCNN (Synthetic)  & $77.5\pm22.3$ & $93.7\pm10.9$ & $97.1\pm0.7$ & $84.4\pm13.0$ & $89.1\pm16.6$ & $\mathbf{95.6}\pm1.1$ & $76.2\pm10.6$ & $90.6\pm6.2$  \\
  
VN (Synthteic)  & $65.5\pm18.7$ & $93.7\pm4.9$ & $80.7\pm17.6$ & $59.3\pm11.0$ & $92.5\pm4.9$ & $82.5\pm15.0$ & $77.4\pm6.1$ & $62.0\pm12.9$
 
% \hline

% Geometric & $81.8\pm15.1$ & $90.6\pm6.1$ & $92.0\pm15.0$ & $94.6\pm3.7$ & $95.0\pm4.5$ & $88.3\pm6.1$ & $77.6\pm11.3$ & $88.3\pm15.4$
    
\vspace{-7pt}
\end{tabular}}
\caption{One-shot generalization on real-world scans from the Dynlab dataset \citep{huang2021multibodysync}.\vspace{-5pt}}
\label{tab:dynlab}
\end{table}

\section{Related work}
\paragraph{Global Equivariance.}
We introduce a novel method for piecewise $E(3)$ equivariance in point networks. 
Euclidean group symmetry has been studied in point networks mainly in describing architectures that accommodate global transformations \citep{chen2019clusternet,thomas2018tensor, fuchs2020se,chen2021equivariant,deng2021vector,assaad2023vntransformer,zisling2022vntnet,katzir2022shape,poulenard2021functional,puny2022frame}. These was shown to perform well in various applications including reconstruction \citep{deng2021vector,chatzipantazis2022se,chen20223d}, pose estimation \citep{li2021leveraging,lin2023coarse,pan2022so,sajnani2022condor,zhu2022correspondence}, and robot manipulation~\citep{simeonov2022neural,higuera2023neural,xue2023useek} tasks. Some works have dealt with respecting the symmetry by manipulating their input representation \citep{deng2018ppf,zhang2019rotation,gojcic2019perfect}. A popular line of work utilizes the theory of spherical harmonics to achieve equivariance \citep{worrall2017harmonic,esteves2018learning,liu2018deep,weiler20183d,cohen2018spherical}. 
%Baking the symmetry into the network architecture with high order SO(3) representations as features was shown to have maximal expressivity \citep{thomas2018tensor, fuchs2020se, romero2020group, dym2020universality}. 
% The VN framework~\citep{deng2021vector} %for building such a network was constructed by lifting the activation of various network layers to 3-space and allowing them to transform along with the input. This has been 
% was extended to transformer architecture in \cite{assaad2023vntransformer, zisling2022vntnet}, where in \cite{chen20223d} a local graph structure and scale equivariance was introduced to further boost reconstruction details. In this work, we rely on the frame averaging framework~\citep{puny2022frame}, which achieves equivariance by averaging over the group operation. 
 \vspace{-10pt}
\paragraph{Object-Level and Part-Based Equivariance}
Several works have studied the equivariance of parts. 
EON~\citep{yu2022eon} and EFEM~\citep{lei2023efem} both studied object-level equivariance in scenes. EON used a manually tuned `suspension' to compute an equivariant object frame in which the context is aggregated. In EFEM, instance segmentation is achieved by training a shape prior using a shape collection, and employing it to refine scene regions. Instead, we do not assume prior knowledge of the underlying partition. Equivariance for per-part pose estimation in articulated shape was devised in \cite{liu2023selfsupervised}. Yet their self-supervised approach relies on part grouping according to features that are invariant to \textit{global} rotations which may result in unknown errors when local transformations are introduced. Part-based equivariance was also studied for segmentation in \cite{deng2023banana}, relying on an intriguing fixed-point convergence procedure.

\section{Conclusion}
\vspace{-5pt}
We presented \ShortName, a point network design for approximately piecewise $E(3)$ equivariant models. We implemented \ShortName networks to tackle recognition tasks such as point cloud segmentation, and classification, demonstrating superior generalization over common baselines. On the theoretical side, our work lays the ground for an analysis of piecewise equivariant networks in terms of their equivariance approximation error. The bounds we present in this study serve as merely initial insights on the possibility of controlling the equivariance approximation error, % highlighting the importance of a network design that explicitly aims at controlling the equivariance approximation error. 
and further analysis of our suggested bounds is marked as an interesting future work. Further extending this framework for other 3D tasks, \eg, generative modeling and reconstruction is another interesting research venue. 

\section*{Acknowledgments}
The authors would like to thank Jonah Philion for the insightful discussions and valuable comments. Or Litany is a Taub fellow and is supported by the Azrieli Foundation Early Career Faculty Fellowship.

% We presented a novel neural network architecture for point clouds which is capable of producing piecewise $E(3)$ invariant and equivariant predictions from data without a priori knowledge of partitioning of points. We apply our architecture to various 3D recognition tasks such as point cloud part and instance segmentation, and classification, demonstrating superior generalization over baseline approaches to unseen part classes and orientations. We further demonstrate the effectiveness of our method in a single-shot setting where we can learn to segment a single scene and generalize to different scenes consisting of similar objects. We remark that our method is still limited for the case of extreme partial inputs (e.g. scans consisting only of small parts of objects). In the future, we would to address this limitation as well as explore piecewise $E(3)$ symmetric networks in domains beyond point clouds, such as neural fields.


\bibliography{iclr2024_conference}
\bibliographystyle{iclr2024_conference}

\appendix
\section{Appendix}

\subsection{Proofs}
\subsubsection{Proof of Lemma \ref{lm:basic}}
\begin{proof}(Lemma \ref{lm:basic})
Let $\mX \in U$, $\mZ \in \{0,1\}^{n\times k}$, and $g\in G$. Then,
\begin{align*}
    \psi(g\cdot(\mX,\mZ), \mZ) &= \sum_{j=1}^k \psi_b(g\cdot(\mX,\mZ) \odot \mZ\ve_j\vone_d^T) \odot \mZ \ve_j\vone^T = \\
    \sum_{j=1}^k \psi_b(\sum_{j=1}^{k} \left(g_j\cdot \mX\right) \odot (\mZ \ve_j \one_d^T)) \odot \mZ \ve_j\vone^T &= \sum_{j=1}^k g_j \cdot \psi_b( \mX \odot \mZ \ve_j \one_d^T) \odot \mZ \ve_j\vone^T
\end{align*}
where the last equality follows from the fact the $\psi_b$ is $E(d)$ equivariant and the second equality from the fact that $\mZ\ve_j \odot \mZ\ve_j' = \vzero$ for $j\neq j'$. Lastly, for any permutation $\sigma_k(\cdot)$, we have, 
$$
\sum_{j=1}^k \psi_b(\mX \odot \mZ\ve_{\sigma_k(j)}\vone_d^T) \odot \mZ \ve_{\sigma_k(j)}\vone^T = \sum_{j=1}^k \psi_b(\mX \odot \mZ\ve_{j)}\vone_d^T) \odot \mZ \ve_j\vone^T
$$

\end{proof}

\subsubsection{Proof of Theorem \ref{thm:main}}
\begin{proof}(Theorem \ref{thm:main})
Let $\phi:U \rightarrow U'$ be of the form 
    \begin{equation}
      \phi(\mX) = \sum_{j=1}^k \psi_b(\mX \odot \mZ_{*} \ve_j \vone^T_d)\odot \mZ_{*} \ve_j \vone^T,  
    \end{equation}
    
    where $(\mZ_{*})_{i,:} = \ve_{\argmax_{j} Q(\mZ \vert \mX)_{ij}}$, and $\psi_b : U \rightarrow U'$ is an $E(d)$ equivariant backbone.

    Let $A = \left\{\mZ \neq \mZ_{*}\right\}$. Then,
    $$
    Q(A) \leq \sum_{i=1}^n (1 -  Q(\ve_i \mZ =\ve_i \mZ_{*})) = \sum_{i=1}^n (1 -  Q_{ij(i)_*})
    $$
    where $j(i)_* = \argmax_j Q_{ij}$. Then, we set
    $$
    \delta (Q) = \sum_{i=1}^n (1 -  Q_{ij(i)_*}).
    $$
    Clearly $\delta$ satisfies conditon \ref{eq:delta}. Now, Let $Q$ satisfying condition \ref{eq:lambda} w.r.t. $\lambda$.
    Let $B = \left\{\mZ \vert \ \exists \ 1\leq i,j \leq n    \text{ s.t. } \left(\mZ  \mZ^T\right)_{ij} > \left(\widehat{\mZ}  \widehat{\mZ}^T\right)_{ij}  \right\}$.
    Then,
    $$
    \left\{\mZ \right\} = (B \cap A) \cup (B \cap A^C) \cup (B^C\cap A) \cup (B^C \cap A^C ).
    $$

    Note that for $B^C \cap A^C $ there is no equivariance approximation error. For $(B \cap A)$, and $(B^C \cap A)$ we can bound using $\delta(Q)$. Lastly, $\mZ \in (B \cap A^C)$ means $\mZ_*$ is a "bad" partition, thus $\lambda(Q)\leq \lambda(Q_\textrm{simple})$. To conclude, we use a union bound composed of the decomposition above to get that,
    $$
    \mathbb{E}_{Q_{\mZ|\mX}} \norm{\phi\left( g\cdot (\mX,\mZ) \right) - g\cdot (\phi (\mX),\mZ)} \leq (\lambda(Q_{\textrm{simple}}) + \delta(Q))M.
    $$
\end{proof}

\subsection{Q Prediction}
% \end{algorithm}
\begin{wrapfigure}[14]{r}{0.33\textwidth}
  \begin{center}
  \vspace{-18pt}
    \includegraphics[width=0.33\textwidth]{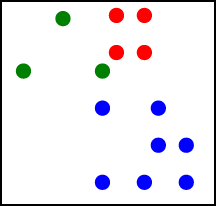}
  \vspace{-20pt}
  \end{center}
  \caption{2D toy example consisting of $n=14$ points, partitioned into $3$ parts.}
  \label{fig:inset_2d_toy}
\end{wrapfigure}

\begin{figure}[h]
    \centering
    \includegraphics[width=0.8\linewidth]{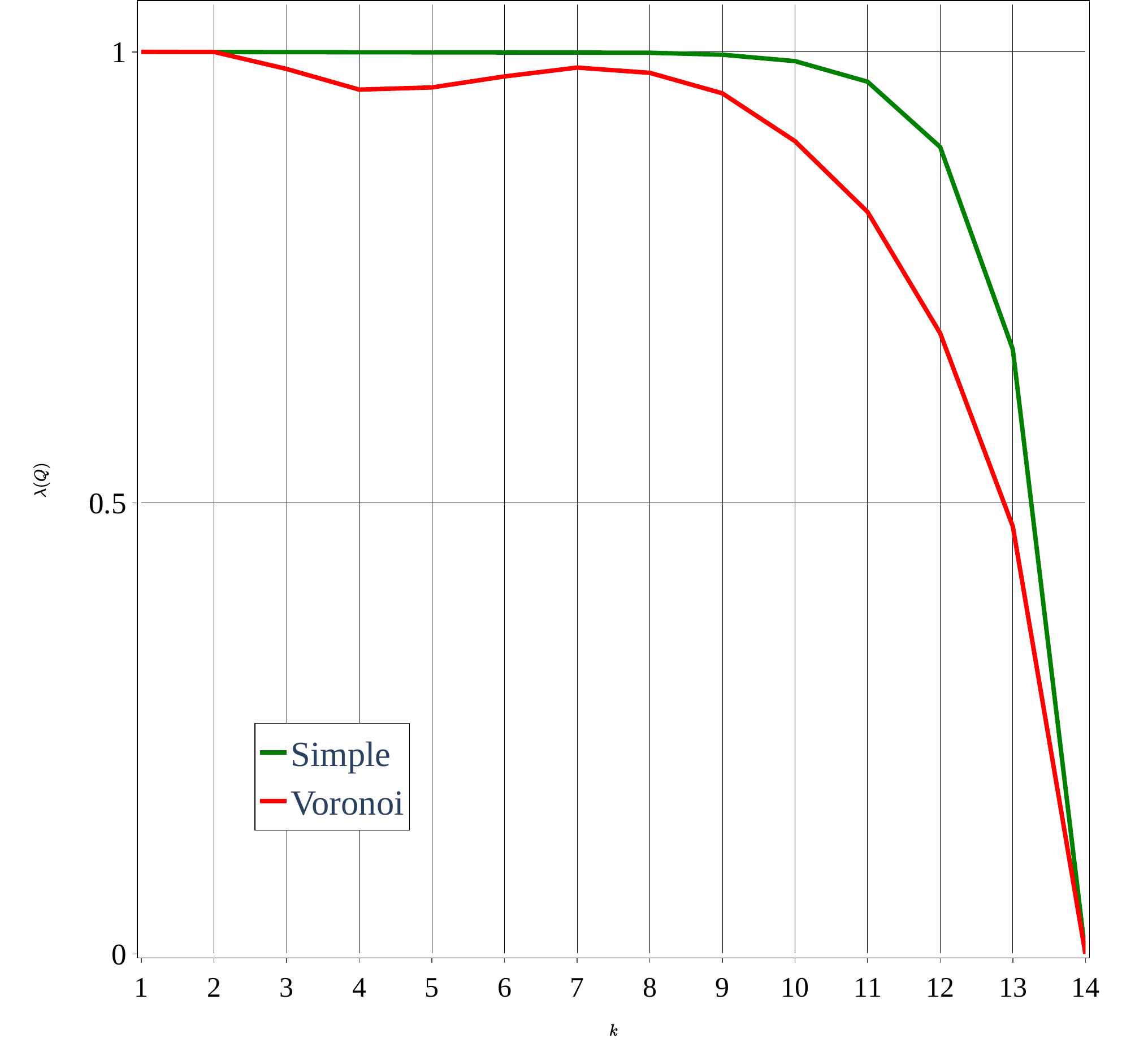}
    \vspace{-2em}
    \caption{\rev{The probability of drawing a bad partition, $\lambda(Q_{\text{simple}})$, as $k\rightarrow n$, for a 2D toy example with $n=14$ points.}}
    \label{fig:lambda_q}
\end{figure}

In this section we provide an empirical validation to the expected behavior of $\lambda(Q_\text{simple})$ as $k\rightarrow n$. To that end, we examine a 2D toy example, featuring $n=14$ points partitioned to $3$ groups. Figure \ref{fig:inset_2d_toy} shows this toy example, with distinct colors denoting the ground truth partition. Figure \ref{fig:lambda_q} shows a plot of $\lambda(Q)$ values for $k \in [1,14]$. The green line shows $\lambda(Q)$ for the simple $Q$ model, defined by a uniform draw of $k$ parts partition, where each part includes at least one point. The red line shows $\lambda(Q)$ for a $Q$ model, defined by a Voronoi partition with centers drawn randomly proportionally to $k$ furthest point sampling. Note that as expected, $\lambda(Q) \rightarrow 0$ as $k\rightarrow n$.
% \begin{wrapfigure}[10]{r}{0.5\textwidth}
%   \begin{center}
%   \vspace{-15pt}
%     \includegraphics[width=0.48\textwidth]{figs/lambda.pdf}
%   % \vspace{20pt}
%   \end{center}
%   \label{fig:inset_intro}
% \end{wrapfigure}

Next, we provide in Alg.~\ref{alg:qpred} a detailed description of our $Q$ prediction algorithm.

\begin{algorithm}[h]
    \caption{Q prediction}\label{alg:qpred}
    
     \textbf{Input: } $\mY$; $\tau > 0$ merge threshold and $f$ merge frequency\\
    
    \begin{algorithmic}
    %\Procedure{MyProcedure}{}
%    \Procedure{MyProcedure}{$x,y$}
%     % Input:
%     \Comment{Input: x}
%     % Output:
%     \Comment{Output:y}
        %$(\mu_j) \gets \text{random sample of k points from \phi(\mX)}$
        %\State $i$ \gets $0$
        \State $i \gets 0 $
        \State $(\vmu_j) \gets \textit{random furthest point sample of } k \textit{ points from } \mY$
        \State $\pi_j \gets \frac{1}{k} $

        \While{ $i < $ \textit{max iter}}\\
            \State $\gamma_{ij} \gets \frac{\pi_j \mathcal{N}(\mY_i;\vmu_j)}{\sum_l \pi_l \mathcal{N}(\mY_i;\vmu_l)}$

            \State $\vmu_j \gets \sum_i \frac{\gamma_{ij}}{\sum_{i'} \gamma_{i'j}}\mY_i$
            \State $\pi_j \gets \frac{\sum_i \gamma_{ij}}{n}$
            \If {$i \textit{ mod } f == 0$}
                \State $(j,j') \gets \argmin\limits_{\left\{j,j'\right\} \in \left\{j|\pi_j > 0\right\}} \KL(\mathcal{N}(\cdot;\vmu_j)\vert \vert \mathcal{N}(\cdot;\vmu_j'))$
                \State $ d \gets \KL(\mathcal{N}(\cdot;\vmu_j)\vert \vert \mathcal{N}(\cdot;\vmu_j'))$
                \While { $d < \tau$}
                    
                    \State $\pi_j \gets \pi_j + \pi_j'$
                    \State $\pi_j' \gets 0$
                    \State $(j,j') \gets \argmin\limits_{\left\{j,j'\right\} \in \left\{j|\pi_j > 0\right\}} \KL(\mathcal{N}(\cdot;\vmu_j)\vert \vert \mathcal{N}(\cdot;\vmu_j'))$
                    \State $ d \gets \KL(\mathcal{N}(\cdot;\vmu_j)\vert \vert \mathcal{N}(\cdot;\vmu_j'))$
            \EndWhile
            \EndIf
            
            \State $i \gets i + 1 $
        \EndWhile
        
        \State $ (\tilde{\vmu}_j,\tilde{\pi}_j) \gets (\vmu_j,\pi_j)$
        \State $ (\vmu^*_j,\pi^*_j) \gets (\tilde{\vmu}_j,\tilde{\pi}_j) + I^{-1}(\tilde{\vmu}_j,\tilde{\pi}_j)s(\mY;(\tilde{\vmu}_j,\tilde{\pi}_j))$        
        \State $Q^{\textrm{pred}}_{ij} \gets \frac{\mathcal{N}(\vy_i;\vmu^*_j,\sigma)\pi^*_j}{\sum_{j=1}^{k}\mathcal{N}(\vy_i;\vmu^*_j,\sigma)\pi^*_j}$
    \end{algorithmic}
     \textbf{Output}: $Q^{\textrm{pred}}$, a (differential) minimizer of $E(\mY)$ 
     %\hspace*{\algorithmicindent}
\end{algorithm}

\subsection{Additional Implementation Details}

\subsubsection{Architecture}
We start by describing our concrete construction for the encoder, $\mathtt{e}$ and $\mathtt{d}$ used in our experiments. The network consists of \ShortName layers of the form,

\begin{align*}
\mathrm{\ShortName}(n,a_{\text{in}},b_{\text{in}},a_{\text{out}},b_{\text{out}}) &:\Real ^ {n \times (a_\text{in} + 3\times b_\text{in})} \too \Real ^{ n \times (a_{\text{out}} + 3 \times b_\text{out})} 
\end{align*}

Then, the encoder consists of the following blocks:

\begin{align*}
&\mathrm{\ShortName}(n,0,2,17,5)  \rightarrow \\
&\mathrm{\ShortName}(n,17,5,17,5) \rightarrow \\
&\mathrm{\ShortName}(n,0,2,17,5) \rightarrow \mathrm{\ShortName}(n,0,2,65,21).
\end{align*}

The decoder consists of the following block for the segmentation task:
$$
    \mathrm{\ShortName}(n,65,21,24,0),
$$
and for the classification task:
$$
    \mathrm{\ShortName}(1,65,21,9,0).
$$

Each \ShortName block is built on equivariant backbone, implemented with Frame Averaging. In turn, the backbone symmetrize a pointnet network $\psi$. We now describe its details.

The network consists of layers of the form 
\begin{align*}
\mathrm{FC}(n,d_{\text{in}},d_{\text{out}}):\mX &\mapsto \nu\parr{\mX \mW + \one \vb^T } \\
\mathrm{MaxPool}(n,d_{\text{in}}) : \mX &\mapsto \one [\max{\mX \ve_i}]
\end{align*}
where $\mX \in \Real ^ {n\times d_\text{in}}$, $\mW\in \Real^{d_\text{in}\times d_\text{out}}$, $\vb\in\Real^{d_\text{out}}$ are the learnable parameters, $\one\in\Real^n$ is the vector of all ones, $[\cdot]$ is the concatenation operator, $\ve_i$ is the standard basis in $\Real^{d_\text{in}}$, and $\nu$ is the $\mathrm{ReLU}$ activation. We used the following architecture for the first \ShortName layer:
\begin{align*}
&\mathrm{FC}(n,6,96) \stackrel{L_1}\rightarrow \mathrm{FC}(n,96,128)  \stackrel{L_2}\rightarrow \mathrm{FC}(n,128,160) \stackrel{L_3}\rightarrow  \mathrm{FC}(n,160,192) \stackrel{L_4}\rightarrow \\
&\mathrm{FC}(n,192,224) \stackrel{L_5} \rightarrow \mathrm{MaxPool}(n,224)  \stackrel{L_6}\rightarrow  [L_1,L_2,L_3,L_4,L_5,L_6] \stackrel{L_7} \rightarrow   \\
&\mathrm{FC}(n,1024,256) \stackrel{L_8}\rightarrow \mathrm{FC}(n,256,256)  \stackrel{L_9} \rightarrow \mathrm{FC}(n,128,32).
\end{align*}

For the second and third,
\begin{align*}
&\mathrm{FC}(n,32,96) \stackrel{L_1}\rightarrow \mathrm{FC}(n,96,128)  \stackrel{L_2}\rightarrow \mathrm{FC}(n,128,160) \stackrel{L_3}\rightarrow  \mathrm{FC}(n,160,192) \stackrel{L_4}\rightarrow \\
&\mathrm{FC}(n,192,224) \stackrel{L_5} \rightarrow \mathrm{MaxPool}(n,224)  \stackrel{L_6}\rightarrow  [L_1,L_2,L_3,L_4,L_5,L_6] \stackrel{L_7} \rightarrow   \\
&\mathrm{FC}(n,1024,256) \stackrel{L_8}\rightarrow \mathrm{FC}(n,256,256)  \stackrel{L_9} \rightarrow \mathrm{FC}(n,128,32).
\end{align*}

And lastly,
\begin{align*}
&\mathrm{FC}(n,32,96) \stackrel{L_1}\rightarrow \mathrm{FC}(n,96,128)  \stackrel{L_2}\rightarrow \mathrm{FC}(n,128,160) \stackrel{L_3}\rightarrow  \mathrm{FC}(n,160,192) \stackrel{L_4}\rightarrow \\
&\mathrm{FC}(n,192,224) \stackrel{L_5} \rightarrow \mathrm{MaxPool}(n,224)  \stackrel{L_6}\rightarrow  [L_1,L_2,L_3,L_4,L_5,L_6] \stackrel{L_7} \rightarrow   \\
&\mathrm{FC}(n,1024,256) \stackrel{L_8}\rightarrow \mathrm{FC}(n,256,256)  \stackrel{L_9} \rightarrow \mathrm{FC}(n,128,128).
\end{align*}

\subsubsection{Hyper parameters and training details}
We set $\sigma_l = (0.002,0.005,0.008,0.1)$. The number of iterations for the EM was 16. 
We trained our networks using the \textsc{Adam} \citep{kingma2014adam} optimizer, setting the batch size to $8$. We set a fixed learning rate of $0.001$. All models were trained for $3000$ epochs. Training was done on a single Nvidia V-100 GPU, using \textsc{pytorch} deep learning framework \citep{paszke2019pytorch}.

\subsection{Additional Results}
\begin{figure}[t]
    \centering
    \includegraphics[width=\linewidth]{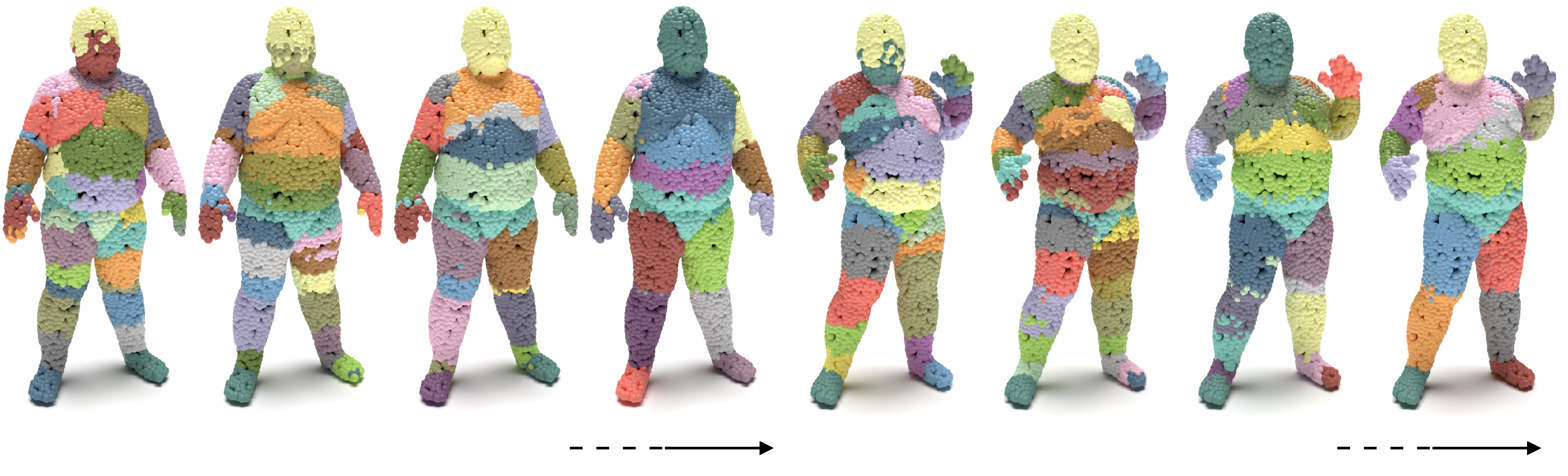}
    \vspace{-2em}
    \caption{\rev{\ShortName encoder's learned partitions, $Q^\text{pred}$, extracted from two test-set examples in the human body segmentation experiment. In each group of $4$ elements, the leftmost column shows $Q^{\text{pred}}$ partitions, with subsequent layers' partitions ordered left-to-right, culminating in the rightmost column that shows the encoder's last layer partition.  }}
    \label{fig:part_human}
\end{figure}
\begin{figure}[t]
    \centering
    \includegraphics[width=\linewidth]{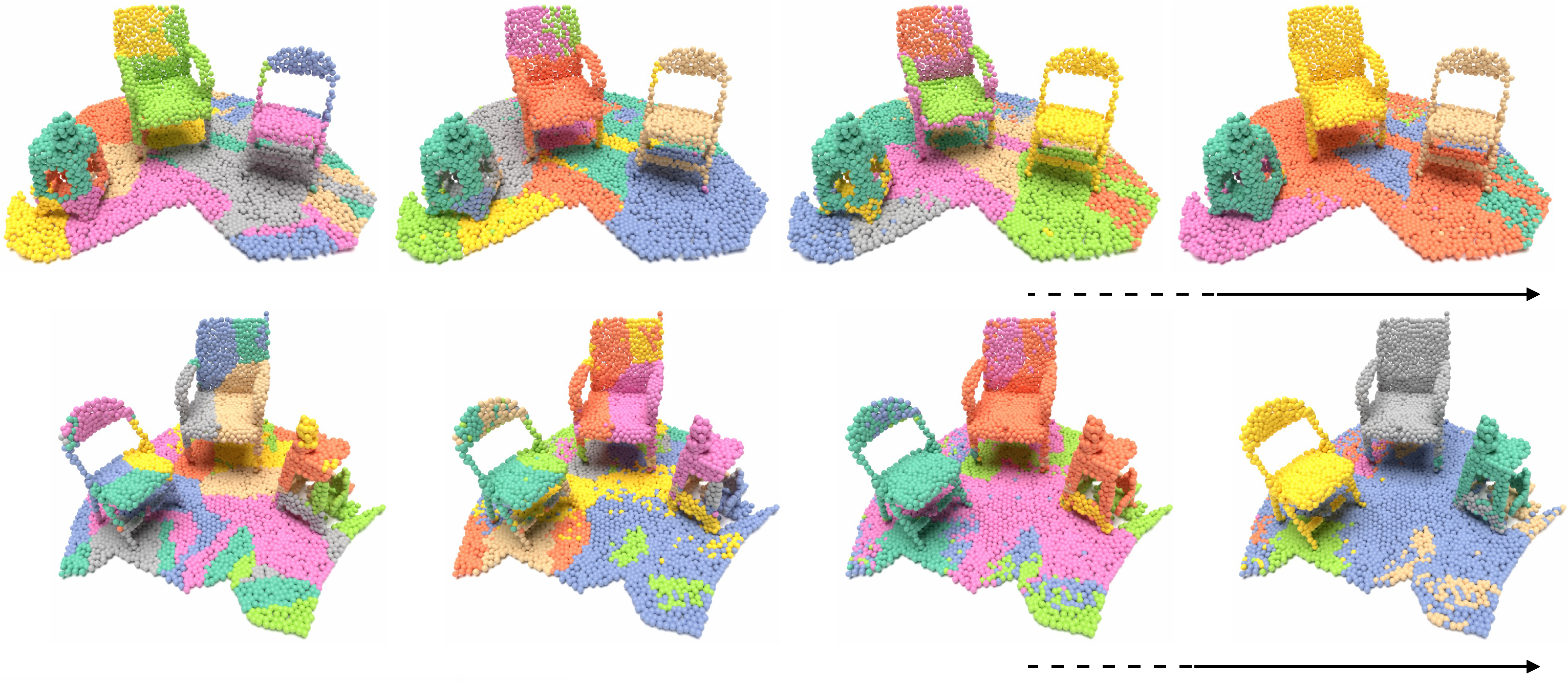}
    \vspace{-2em}
    \caption{\rev{\ShortName encoder's learned partitions, $Q^\text{pred}$, extracted from the one shot segementation experiment. In the top row, layer partitions of a single training example are shown, while the bottom row shows layer partitions of an unseen test example. The leftmost column shows $Q^{\text{pred}}$ partitions, with subsequent layers' partitions ordered left-to-right, culminating in the rightmost column that shows the encoder's last layer partition.  }}
    \label{fig:part_dyn}
\end{figure}

\rev{In this section, we present visualizations of the learned partitions $Q^{\text{pred}}$ across layers in the \ShortName encoder. Figure \ref{fig:part_human} shows the learned \ShortName encoder layers partitions from the experiment in section \ref{ss:humans}, while Figure \ref{fig:part_dyn} shows partitions from the experiment in section \ref{ss:rooms}. Each input point is assigned distinctive colors according to $\argmax_j Q^{\text{pred}}_{ij}$. It is worth noting that progressing from left to right, the predicted partitions tend to become coarser, a behavior encouraged by setting the hyper-parameter $\sigma_{l+1} > \sigma_l$.   }

\subsection{Subject Classification Experiment}

\begin{table}[h!]
\centering
\resizebox{0.5\textwidth}{!}{%
\begin{tabular}{l|llllllll}
Method &  PointNet & DGCNN & VN & Ours
     \\ 
     \hline
Accuracy (\%)  & $18.5$ & $32.1$ & $28.2$ & $71.4$ 
\vspace{-7pt}
\end{tabular}}
\caption{Subject classification accuracy comparison.\vspace{-5pt}}
\label{tab:cls}
\end{table}

\begin{figure}[h!]
    \centering
    \includegraphics[width=\linewidth]{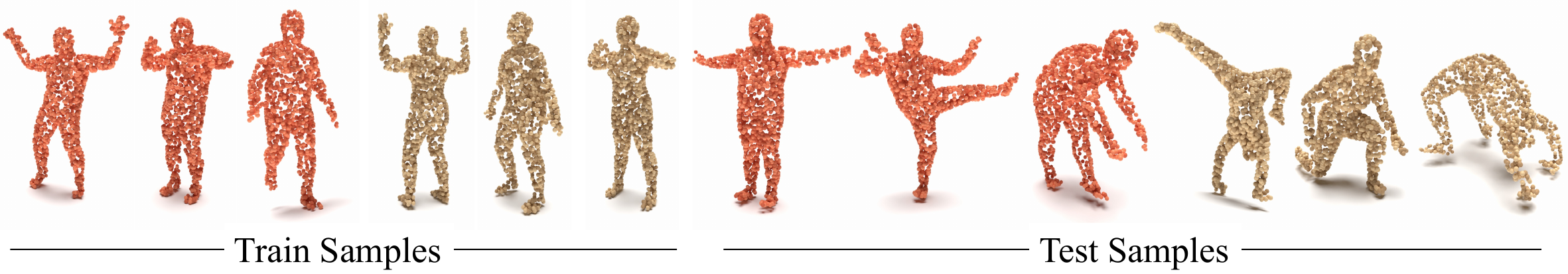}
    \vspace{-2em}
    \caption{Training and test set visualization for the subject classification task.}
    \label{fig:class}
\end{figure}

Here we provide the results of the point cloud classification experiment described in the main text. Fig.~\ref{fig:class} shows several typical examples from the considered split. Note the relatively large difference in the distribution of poses. Tab.~\ref{tab:cls} logs the quantitative evaluation, validating our framework's superiority in this case as well.

% \subsection{Limitations}

% \jia{Discuss limitations here.}
% \begin{wrapfigure}[4]{r}{0.33\textwidth}
%   \begin{center}
%   \vspace{-10pt}
%     \includegraphics[width=0.33\textwidth]{figs/failure.pdf}
%   \end{center}
%   \label{fig:failure}
% \end{wrapfigure}

\end{document}